\documentclass[conference]{IEEEtran}
\usepackage{times}
\usepackage{algorithm}
\usepackage{algorithmic}
\usepackage{amssymb}
\usepackage{amsmath}
\usepackage{amsfonts}
\usepackage{color}
\let\labelindent\relax

\usepackage{enumitem}

\usepackage{graphicx} 
\usepackage{subfigure}

\usepackage{flushend}

\usepackage[numbers]{natbib}
\usepackage{multicol}
\usepackage{url}
\usepackage[bookmarks=true]{hyperref}

\usepackage{etoolbox}
\makeatletter
\patchcmd{\@makecaption}
  {\scshape}
  {}
  {}
  {}
\patchcmd{\@makecaption}
  {\\}
  {.\ }
  {}
  {}
\makeatother

\begin{document}
\title{One-Shot Imitation from Observing Humans\\ via Domain-Adaptive Meta-Learning}

\author{\authorblockN{Tianhe Yu*, Chelsea Finn*, Annie Xie, Sudeep Dasari, Tianhao Zhang, Pieter Abbeel, Sergey Levine}
\authorblockA{University of California, Berkeley\\
Email: \{tianhe.yu,cbfinn,anniexie,sdasari,tianhao.z,pabbeel,svlevine\}@berkeley.edu\\
* denotes equal contribution}}


%

\maketitle

\begin{abstract}
Humans and animals are capable of learning a new behavior by observing others perform the skill just once. We consider the problem of allowing a robot to do the same -- learning from a raw video pixels of a human, even when there is substantial domain shift in the perspective, environment, and embodiment between the robot and the observed human. Prior approaches to this problem have hand-specified how human and robot actions correspond and often relied on explicit human pose detection systems. In this work, we present an approach for one-shot learning from a video of a human by using human and robot demonstration data from a variety of previous tasks to build up prior knowledge through meta-learning. Then, combining this prior knowledge and only a single video demonstration from a human, the robot can perform the task that the human demonstrated. We show experiments on both a PR2 arm and a Sawyer arm, demonstrating that after meta-learning, the robot can learn to place, push, and pick-and-place new objects using just one video of a human performing the manipulation.
\end{abstract}

\IEEEpeerreviewmaketitle

\section{Introduction}
\label{sec:intro}

Demonstrations provide a descriptive medium for specifying robotic tasks. Prior work has shown that robots can acquire a range of complex skills through demonstration, such as table tennis~\cite{table_tennis}, lane following~\cite{alvinn}, pouring water~\cite{pastor}, drawer opening~\cite{drawer_opening}, and multi-stage manipulation tasks~\cite{vr_imitation}.
However, the most effective methods for robot imitation differ significantly from how humans and animals might imitate behaviors: while robots typically need to receive demonstrations in the form of kinesthetic teaching~\cite{pastor_kinesthetic,akgun2012trajectories} or teleoperation~\cite{hapticteleop_lfd,florida_vision,vr_imitation}, humans and animals can acquire the gist of a behavior simply by \emph{watching} someone else. In fact, we can adapt to variations in morphology, context, and task details effortlessly, compensating for whatever \emph{domain shift} may be present and recovering a skill that we can use in new situations~\cite{correspondence_problem2}. Additionally, we can do this from a very small number of demonstrations, often only one.
How can we endow robots with the same ability to learn behaviors from raw third person observations of human demonstrators?

Acquiring skills from raw camera observations presents two major challenges. First, the difference in appearance and morphology of the human demonstrator from the robot introduces a systematic domain shift, namely the correspondence problem~\cite{correspondence_problem1,correspondence_problem2}.
Second, learning from raw visual inputs typically requires a substantial amount of data,
with modern deep learning vision systems using hundreds of thousands to millions of images~\cite{xiang2017posecnn,kim2017satellite}.
In this paper, we demonstrate that we can begin to address both of these challenges through a single approach based on meta-learning.
Instead of manually specifying the correspondence between human and robot,
which can be particularly complex for skills where different morphologies require different strategies,
we propose a data-driven approach.
Our approach can acquire new skills from only one video of a human. To enable this, it builds a rich prior over tasks during a \emph{meta-training} phase, where both human demonstrations and teleoperated demonstrations are available for a variety of other, structurally similar tasks.
In essence, the robot learns how to learn from humans using this data. After the meta-training phase, the robot can acquire new skills by combining its learned prior knowledge with one video of a human performing the new skill.

\begin{figure}
\setlength{\unitlength}{0.5\columnwidth}
\begin{picture}(1.0,1.23) \linethickness{0.5pt}
\put(-0.02,0.0){\includegraphics[width=1.03\columnwidth]{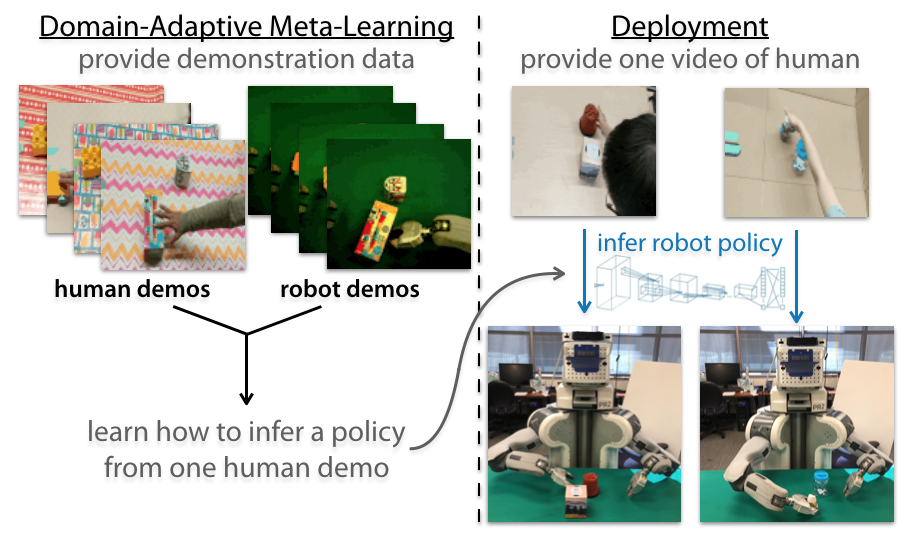}}
\end{picture}
\vspace{-0.38cm}
\caption{
\label{fig:teaser}
\!\!\!After meta-learning with human and robot demonstration data, the robot learns to recognize and push a new object from one video of a human.}
\vspace{-0.24in}
\end{figure}

The main contribution of this paper is a system for learning robotic manipulation skills from a single video of a human by leveraging large amounts of prior meta-training data, collected for different tasks.
When deployed, the robot can adapt to a particular task with novel objects using just a single video of a human performing the task with those objects  (e.g., see Figure~\ref{fig:teaser}). The video of the human need not be from the same perspective as the robot, or even be in the same room. The robot is trained using videos of humans performing tasks with various objects along with demonstrations of the robot performing the same task.
Our experiments on two real robotic platforms demonstrate the ability to learn directly from RGB videos of humans, and to handle novel objects, novel humans, and videos of humans in novel scenes. 
Videos of the results can be found on the supplementary website.\footnote{\vspace{-0.2cm} The video is available at \url{https://sites.google.com/view/daml}}

\vspace{-0.1cm}
\section{Related Work}
\label{sec:related_work}
\vspace{-0.1cm}

Most imitation learning and learning from demonstration methods operate at the level of configuration-space trajectories~\cite{imitation_survey,lfd_survey}, which are typically collected using kinesthetic teaching~\cite{pastor_kinesthetic,akgun2012trajectories}, teleoperation~\cite{hapticteleop_lfd,florida_vision,vr_imitation}, or sensors on the demonstrator~\cite{human_sensor3,human_sensor2,human_sensor1,kruger2010learning}. Instead, can we allow robots to imitate just by watching the demonstrator perform the task?
We focus on this problem of learning from one video demonstration of a human performing a task, in combination with human and robot demonstration data collected on other tasks.
Prior work has proposed to resolve the correspondence problem
by hand, for example, by manually specifying how human grasp poses correspond to robot grasps~\cite{kragic_grasp} or by manually defining how human activities or commands translate into robot actions~\cite{maryland,lee2013syntactic,rnn_translation,ramirez2015transferring,episodic_memory}. By utilizing
demonstration data of how humans and robots perform each task, our approach learns the correspondence between the human and robot implicitly. Many prior approaches also use explicit hand-tracking systems with carefully engineered pipelines for visual activity recognition~\cite{future_regression,ramirez2015transferring}. In contrast to such approaches, which rely on precise hand detection and a pre-built vision system, our approach is trained end-to-end, seeking to extract the aspects of the human's activity that are the most relevant for choosing actions.
This places less demand on the vision system, requiring it only to implicitly deduce the task and how to accomplish it, rather than precisely tracking the human's body and nearby objects.

Other prior approaches have sought to solve the problem of learning from human demonstrations by explicitly determining the goal
or reward underlying the human behavior (e.g. through inverse reinforcement learning), and then optimizing the reward through reinforcement learning (RL). For example,~\citet{firstperson_forecasting} and~\citet{aus_prediction} learn a model that predicts the outcome of the human's demonstration from a particular scene.
Similarly, other works have learned a reward function based on human demonstrations~\cite{pierre,third_person,abhishek,pierre_time,tai2017socially}. Once the system has learned about the reward function or desired outcome underlying the given task, the robot runs some form of reinforcement learning to maximize the reward or to reach the desired outcome.
This optimization typically requires substantial experience to be collected using the robot for each individual task.
Other approaches assume a known model and perform trajectory optimization to reach the inferred goal~\cite{muelling2015autonomy}.
Because all of these methods consider single tasks in isolation, they often require multiple human demonstrations of the new task (though not all, e.g.~\cite{pierre_time,muelling2015autonomy}).
Our method only requires one demonstration of the new task setting and, at test time, does not require additional experience on the robot nor a known model. And, crucially, all of the data used in our approach is amortized across tasks, such that the amount of data needed for any given individual task as quite small. In contrast, these prior reward-learning methods only handle the single-task setting, where a considerable amount of data must be collected for an individual task.

Instead of using the previously mentioned approaches, we use a meta-learning approach~\cite{thrun,schmidhuber1987}, which in recent works has shown the ability to learn from just one demonstration~\cite{mil,rocky} using prior knowledge built up from many demonstrations of other tasks.
In particular, we extend the approach of model-agnostic meta-learning~\cite{maml,mil}.
Up until now, these approaches have not considered the problem of domain shift between the demonstration used for training and for testing, e.g. learning from videos of humans.

Handling domain shift is a key aspect of this problem, with a shift in both the visual scene and the embodiment of the human/robot, including the degrees of freedom and the physics. Domain adaptation has received significant interest within the machine learning community, especially for varying visual domains~\cite{aytar2011tabula,patel2015visual} and visual shift between simulation and reality~\cite{saenko_plat_grasping,sadeghi2016cad}. Many of these techniques aim to find a representation that is invariant to the domain~\cite{fernando2013unsupervised,gong2013connecting,tzeng2014deep,sadeghi2016cad,domain_gen}. Other approaches have sought to map datapoints from one domain to another~\cite{shrivastava2017learning,yoo2016pixel,bousmalis2016unsupervised,you2017virtual}. The human imitation problem may involve developing invariances, for example, to the background or lighting conditions of the human and robot's environments. However, the physical correspondence between human and robot does not call for an invariant representation, nor a direct mapping between the domains. In many scenarios, a direct physical correspondence between robot and human poses might not exist. 
Instead, the system must implicitly recognize the goal of the human from the video and determine the appropriate action.

\newcommand{\loss}{\mathcal{L}}
\newcommand{\data}{\mathcal{D}}
\newcommand{\task}{\mathcal{T}}

\vspace{-0.2cm}
\section{Preliminaries}
\vspace{-0.1cm}



Our approach builds upon prior work in learning to learn or meta-learning, in order to learn how to infer a robot policy from just one human demonstration. In particular, we will present an extension of the model-agnostic meta-learning algorithm (MAML)~\cite{maml} that allows for the ability to handle domain shift between the provided data (i.e. a human demo) and the evaluation setting (i.e. the robot's execution), and the ability to learn effectively without labels (i.e., the human's actions). In this section, we will overview the general meta-learning problem and the MAML algorithm.

Meta-learning algorithms optimize for the ability to learn new tasks quickly and efficiently.
To do so, they use data collected across a wide range of meta-training tasks and are evaluated based on their ability to learn new meta-test tasks. 
Meta-learning assumes that the meta-training and meta-test tasks are drawn from some distribution $p(\task)$.
Generally, meta-learning can be viewed as discovering the structure that exists between tasks such that,
when the model is presented with a new task from the meta-test set, it can use the known structure to quickly learn the task.
MAML achieves this by optimizing for a deep network's initial parameter setting such that one or a few steps of gradient descent on a few datapoints leads to effective generalization (referred to as few-shot generalization)~\cite{maml}. Then, after meta-training, the learned parameters are fine-tuned on data from a new task.
This meta-learning process can be formalized as learning a prior over functions, and the fine-tuning process as inference under the learned prior~\cite{hb,grant2018recasting}.

Concretely, consider a supervised learning problem with a loss function denoted as $\loss(\theta, \data)$, where $\theta$ denotes the model parameters and $\data$ denotes the labeled data. 
For a few-shot supervised learning problem, MAML assumes access to a small amount of data for a large number of tasks. During meta-training, a task $\task$ is sampled, along with data from that task, which is randomly partitioned into two sets, $\data^\text{tr}$ and $\data^\text{val}$. We will assume that $\data^\text{tr}$ has $K$ examples.
MAML optimizes for a set of model parameters $\theta$ such that one or a few gradient steps on $\data^\text{tr}$ produces good performance on $\data^\text{val}$. Effectively, MAML optimizes for generalization from $K$ examples. Thus, using $\phi_\task$ to denote the updated parameters, the MAML objective is the following:
\vspace{-0.05cm}
\begin{align*}
    \min_\theta  \sum_\task \loss(\theta-\alpha \nabla_\theta \loss(\theta, \data^\text{tr}_\task), \data^\text{val}_\task) 
    = \min_\theta \sum_\task \loss(\phi_\task, \data^\text{val}_\task).
    \vspace{-0.1cm}
\end{align*}
where $\alpha$ is a step size that can be set as a hyperparameter or learned. Moving forward, we will refer to the inner loss function as the \emph{adaptation objective} and the outer objective as the \emph{meta-objective}.
Subsequently, at meta-test time, $K$ examples from a new, held-out task $\task_\text{test}$ are presented and we can run gradient descent starting from $\theta$ to infer model parameters for the new task:
\vspace{-0.1cm}
$$
\phi_{\task_\text{test}} = \theta - \alpha \nabla_\theta \loss(\theta, \data^{\text{tr}}_{\task_\text{test}}).
\vspace{-0.1cm}
$$
For convenience, we will use only one inner gradient step in the equations. However, using multiple inner gradient steps is straight-forward, and frequently done in practice.

\citet{mil} applied the MAML algorithm to one-shot imitation learning problem, using robot demonstrations collected via teleoperation and a mean-squared error behavioral cloning objective for the loss $\loss$. While this enables learning from one robot demonstration at meta-test time, it does not allow the robot to learn from a raw video of a human or handle domain shift between the demonstration medium and the robot.
Next, we will present our approach for one-shot imitation learning from raw video under domain shift.






\newcommand{\humandata}{\data^h}
\newcommand{\humandemo}{\mathbf{d}^h}
\newcommand{\robotdata}{\data^r}
\newcommand{\robotdemo}{\mathbf{d}^r}
\newcommand{\learnedloss}{\loss_\psi}
\newcommand{\bcloss}{\loss_\text{BC}}
\newcommand{\obs}{\mathbf{o}}
\newcommand{\state}{\mathbf{s}}
\newcommand{\action}{\mathbf{a}}
\newcommand{\method}{DAML}

\vspace{-0.0cm}
\section{Learning from Humans}
\label{sec:method}
\vspace{-0.05cm}

In this section, we will present the problem statement of one-shot imitation learning from humans, introduce our method, and discuss a key aspect of our approach: a learned temporal adaptation objective. 

\vspace{-0.05cm}
\subsection{Problem Overview}
\vspace{-0.1cm}

The problem of learning from human video can be viewed as an inference problem, where the goal is to infer the robot policy parameters $\phi_{\task_i}$ that will accomplish the task $\task_i$ by incorporating prior knowledge with a small amount of evidence, in the form of one human demonstration. In order to effectively learn from just one video of a human, we need a rich prior that encapsulates a visual and physical understanding of the world, what kinds of outcomes the human might want to accomplish, and which actions might allow a robot to bring about that outcome.
We could choose to encode prior knowledge manually, for example by using a pre-defined vision system, a pre-determined set of human objectives, or a known dynamics model. However, this type of manual knowledge encoding is task-specific and time-consuming, and does not benefit from data.
We will instead study how we can learn this prior automatically, using human and robot demonstration data from a variety of tasks.

Formally, we will define a demonstration from a human $\humandemo$ to be a sequence of image observations $\obs_1, ..., \obs_T$, and a robot demonstration $\robotdemo$ to be a sequence of image observations, robot states, and robot actions: $\obs_1, \state_1, \action_1, ..., \obs_T, \state_T, \action_T$. The robot state includes the robot's body configuration, such as joint angles, but does not include object information, which must be inferred from the image. We do not make any assumptions about the similarities or differences between the human and robot observations; they may contain substantial domain shift, e.g. differences in the appearance of the arms, background clutter, and camera viewpoint.

Our approach consists of two phases. First, in the meta-training phase, the goal will be to acquire a prior over policies using both human and robot demonstration data, that can then be used to quickly learn to imitate new tasks with only human demonstrations. For meta-training, we will assume a distribution over tasks $p(\task)$, a set of tasks $\{ \task_i \}$ drawn from $p(\task)$ and, for each task, two small datasets containing several human and robot demonstrations, respectively: $(\humandata_{\task_i}, \robotdata_{\task_i})$. After the meta-training phase, the learned prior can be used in the second phase, when the method is provided with a human demonstration of a new task $\task$ drawn from $p(\task)$. The robot must combine its prior with the new human demonstration to infer policy parameters $\phi_{\task}$ that solve the new task. We will next discuss our approach in detail.

\vspace{-0.05cm}
\subsection{Domain-Adaptive Meta-Learning}
\vspace{-0.1cm}

We develop a domain-adaptive meta-learning method, which will allow us to handle the setting of learning from video demonstrations of humans. While we will extend the MAML algorithm for this purpose, the key idea of our approach is applicable to other meta-learning algorithms.
Like the MAML algorithm, we will learn a set of initial parameters, such that after one or a few steps of gradient descent on just one human demonstration, the model can effectively perform the new task. Thus, the data $\data^\text{tr}_\task$ will contain one human demonstration of task $\task$, and the data $\data^\text{val}_\task$ will contain one or more robot demonstrations of the same task. 

Unfortunately, we cannot use a standard imitation learning loss for the inner adaptation objective computed using $\data^\text{tr}_\task$, since we do not have access to the human's actions. Even if we knew the human's actions, they will typically not correspond directly to the robot's actions.
Instead, we propose to meta-learn an adaptation objective that does not require actions, and instead operates only on the policy activations. The intuition behind meta-learning a loss function is that we can acquire a function that only needs to look at the available inputs (which do not include the actions), and still produce gradients that are suitable for updating the policy parameters so that it can produce effective actions after the gradient update. While this might seem like an impossible task, it is important to remember that the meta-training process still supervises the policy with true robot actions during meta-training. The role of the adaptation loss therefore may be interpreted as simply directing the policy parameter update to modify the policy to pick up on the right visual cues in the scene, so that the meta-trained action output will produce the right actions. We will discuss the particular form of $\loss_\psi$ in the following section. 

During the meta-training phase, we will learn both an initialization $\theta$ and the parameters $\psi$ of the adaptation objective $\loss_\psi$. The parameters $\theta$ and $\psi$ are optimized for choosing actions that match the robot demonstrations in $\data^\text{val}_\task$.
After meta-training, the parameters $\theta$ and $\psi$ are retained, while the data is discarded. A human demonstration $\humandemo$ is provided for a new task $\task$ (but not a robot demonstration). To infer the policy parameters for the new task, we use gradient descent starting from $\theta$ using the learned loss $\loss_\psi$ and one human demonstration $\humandemo$: $\phi_\task = \theta - \alpha \nabla_\theta \learnedloss(\theta, \humandemo)$.

We optimize for task performance during meta-training using a behavioral cloning objective that maximizes the probability of the expert actions in $\data^\text{val}$.
In particular, for a policy parameterized by $\phi$ that outputs a distribution over actions $\pi_\phi(\cdot | \obs, \state)$, the behavioral cloning objective is 
\begin{align*}
\bcloss(\phi, \robotdemo) \!=\! \bcloss(\phi,\! \{\obs_{1:T}, \state_{1:T}, \action_{1:T}\}) \!= \!\! \sum_t \log \pi_\phi(\action_t | \obs_t, \state_t)
\end{align*}
Putting this together with the inner gradient descent adaptation, the meta-training objective is the following:
$$
\min_{\theta, \psi} \sum_{\task \sim p(\task)} \sum_{\humandemo \in \humandata_\task} \sum_{\robotdemo \in \robotdata_\task} \bcloss( \theta - \alpha \nabla_\theta \learnedloss(\theta, \humandemo), \robotdemo). 
$$
The algorithm for optimizing this meta-objective is summarized in Algorithm~\ref{alg:metatrain}, whereas the procedure for learning from humans at meta-test time is shown in Algorithm~\ref{alg:metatest}. 
We will next discuss the form of the learned loss function, $\learnedloss$, which is critical for effective learning.

\begin{algorithm}[t]
    \caption{Meta-imitation learning from humans}
    \label{alg:metatrain}
\begin{algorithmic}
\REQUIRE $\{(\humandata_{\task_i}, \robotdata_{\task_i})\}$: human and robot demonstration data for a set of tasks $\{\task_i\}$ drawn from $p(\task)$
\REQUIRE $\alpha$, $\beta$: inner and outer step size hyperparameters 
\WHILE{training}
    \STATE Sample task $\task \sim p(\task)$ \COMMENT{or minibatch of tasks}
    \STATE Sample video of human $\humandemo \sim \humandata_\task$
    \STATE Compute policy parameters $\phi_\task = \theta - \alpha \nabla_\theta \learnedloss (\theta, \humandemo)$
    \STATE Sample robot demo $\robotdemo \sim \robotdata_\task$
    \STATE Update $(\theta, \psi) \leftarrow (\theta, \psi) - \beta \nabla_{\theta, \psi} \bcloss(\phi_\task, \robotdemo)$
\ENDWHILE
    \STATE Return $\theta, \psi$
\end{algorithmic}
\end{algorithm}

\begin{algorithm}[t]
    \caption{Learning from human video after meta-learning}
    \label{alg:metatest}
\begin{algorithmic}
\REQUIRE meta-learned initial policy parameters $\theta$
\REQUIRE learned adaptation objective $\loss_\psi$
\REQUIRE one video of human demo $\humandemo$ for new task $\task$
\STATE Compute policy parameters $\phi_\task = \theta - \alpha \nabla_\theta \learnedloss (\theta, \humandemo)$
\RETURN $\pi_{\phi}$
\end{algorithmic}
\end{algorithm}

\vspace{-0.05cm}
\subsection{Learned Temporal Adaptation Objectives}
\vspace{-0.1cm}

\begin{figure}[t]
\setlength{\unitlength}{0.5\columnwidth}
\begin{picture}(1.0,0.9) \linethickness{0.5pt}
\put(-0.02,-0.1){\includegraphics[width=1.05\columnwidth]{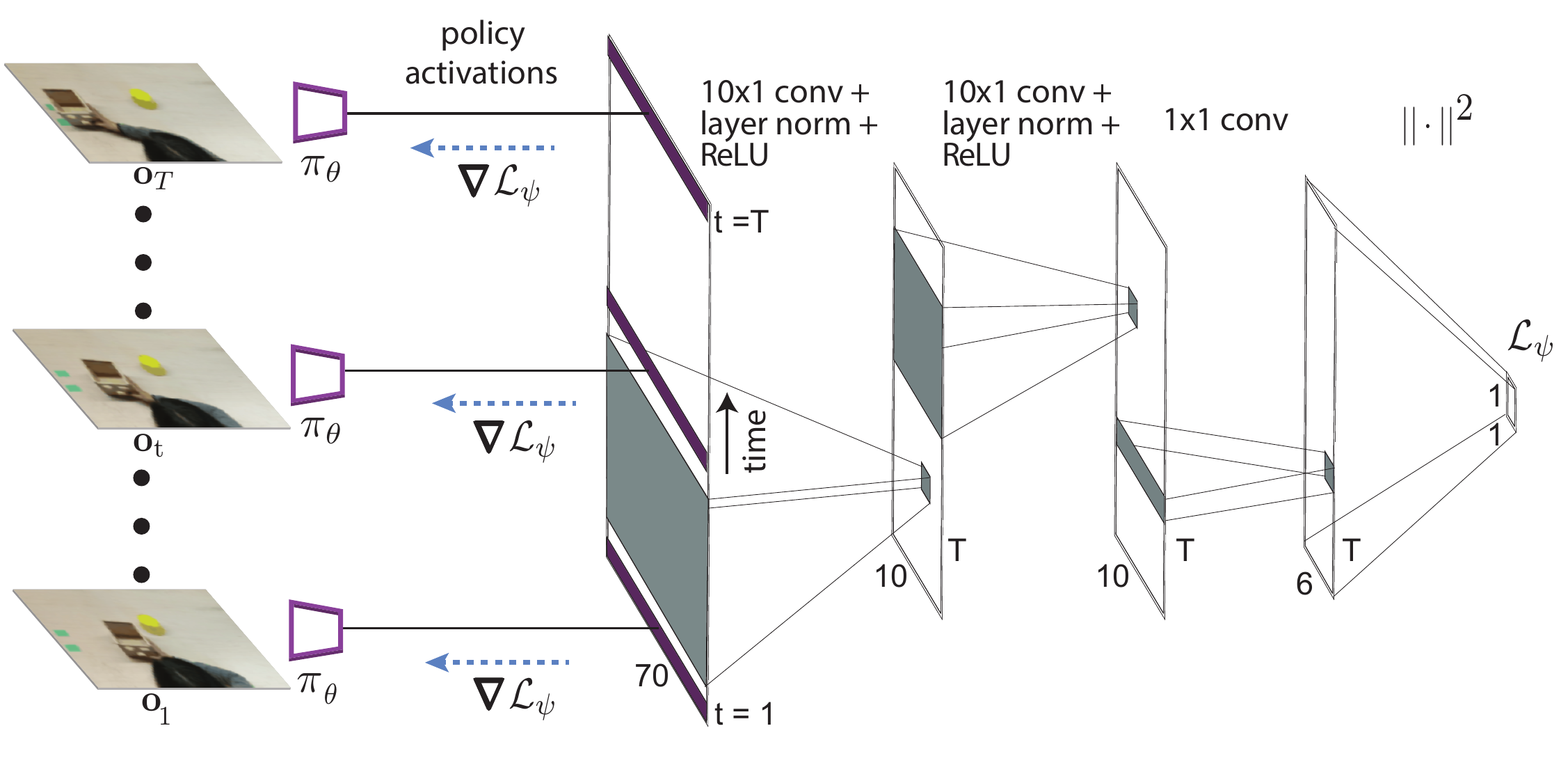}}
\end{picture}
\caption{
\label{fig:learned_loss}
Visualization of the learned adaptation objective, which uses temporal convolutions to integrate temporal information in the video demonstration.}
\vspace{-0.3cm}
\end{figure}

To learn from a video of a human, we need an adaptation objective
that can effectively capture relevant information in the video, such as the intention of the human and the task-relevant objects. While a standard behavior cloning loss is applied to each time step independently, the learned adaptation objective must solve a harder task, since it must provide the policy with suitable gradient information \emph{without} access to the true actions. As discussed previously, this is still possible, since the policy is trained to output good actions during meta-training. The learned loss must simply supply the gradients that are needed to modify the perceptual components of the policy to attend to the right objects in the scene, so that the action output actually performs the right task. However, determining which behavior is being demonstrated and which objects are relevant will often require examining multiple frames at the same time to determine the human's motion. To incorporate this temporal information, our learned adaptation objective therefore couples multiple time steps together, operating on policy activations from multiple time steps.

Since temporal convolutions have been shown to be effective at processing temporal and sequential data~\cite{wavenet}, we choose to adopt a convolutional network to represent the adaptation objective $\loss_\psi$, using multiple layers of 1D convolutions over time. We choose to use temporal convolutions over a more traditional recurrent neural network like an LSTM, since they are simpler and usually more parameter efficient~\cite{wavenet}.
See Figure~\ref{fig:learned_loss} for a visualization.

Prior work introduced a two-head architecture for one-shot imitation, with one head used for the pre-update demonstration and one head used for the post-update policy~\cite{mil}.
The two-head architecture can be interpreted as a learned linear loss function operating on the last hidden layer of the policy network for a particular timestep. The loss and the gradient are then computed by averaging over all timesteps in the demonstration. As discussed previously, a single timestep of an observed video is often not sufficient for learning from video demonstrations without actions.
Thus, this simple averaging scheme is not effective at integrating temporal information. In Section~\ref{sec:experiments}, we show that our learned temporal loss can enable effective learning from demonstrations without actions, substantially outperforming this single-timestep linear loss.

\vspace{-0.05cm}
\subsection{Probabilistic Interpretation}
\vspace{-0.1cm}

\begin{figure}
\setlength{\unitlength}{0.5\columnwidth}
\begin{picture}(1.0,0.78) \linethickness{0.5pt}
\put(0.04,0.0){\includegraphics[width=0.92\columnwidth]{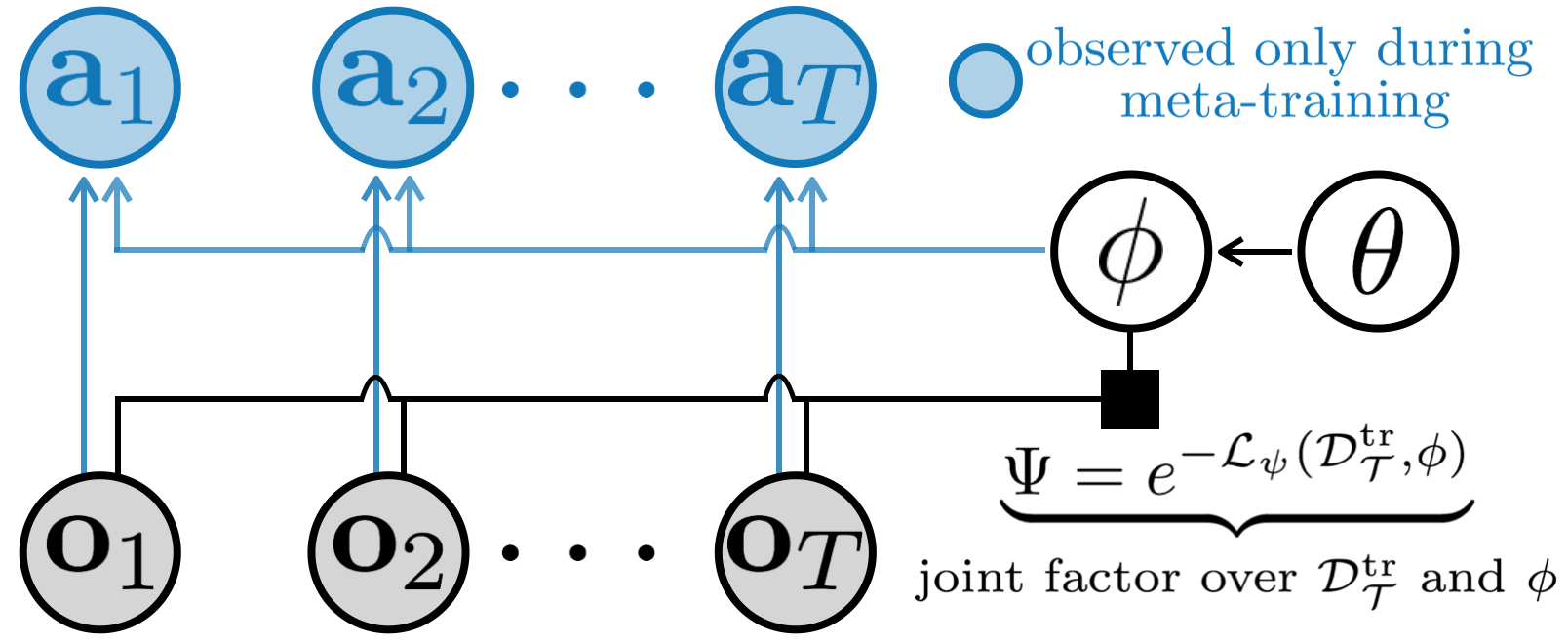}}
\end{picture}
\vspace{-0.3cm}
\caption{
\label{fig:graphical_model}
Graphical model underlying our approach. During meta-training, both the observations $\mathbf{o}_t$ and the actions $\mathbf{a}_t$ are observed, and our method learns $\theta$ and $\Psi$. During meta-testing, only the observations are available, from which our method combines with the learned prior $\theta$ and factor $\Psi$ to infer the task-specific policy parameters $\phi$. }
\vspace{-0.3cm}
\end{figure}

\begin{figure*}
\setlength{\unitlength}{0.5\columnwidth}
\begin{picture}(1.99,1.1) \linethickness{0.5pt}
\put(0.0,0.0){\includegraphics[width=2.05\columnwidth]{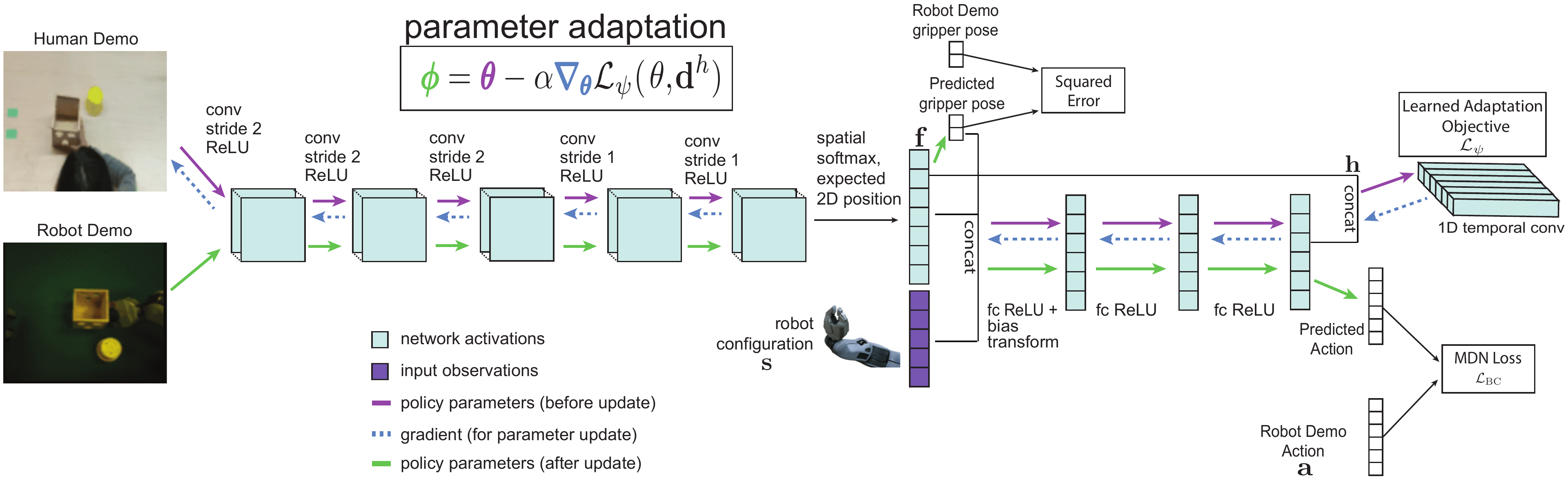}}
\end{picture}
\vspace{-0.4cm}
\caption{
\label{fig:policy}
Illustration of the policy architecture. The policy consists of a sequence of five convolutional (conv) layers, followed by a spatial soft-argmax and fully-connected (fc) layers. The learned adaptation objective $\loss_\psi$ is further illustrated in Figure~\ref{fig:learned_loss}. Best viewed in color.}
\vspace{-0.35cm}
\end{figure*}

One way to interpret meta-learning with learned adaptation objectives is by casting it into the framework of probabilistic graphical models. We can accomplish this by building on a derivation proposed in prior work~\cite{grant2018recasting}, which frames MAML as approximately inferring a posterior over policy parameters $\phi$ given the evidence $\data^\text{tr}_\task = \humandemo_\task$ (the data for a new task $\task$) and a prior over the parameters, given by $\theta$. This prior work shows that a few steps of gradient descent on the likelihood $\log p(\data^\text{tr}_\task | \phi)$ starting from $\phi = \theta$ are approximately equivalent to maximum a posteriori (MAP) inference on $\log p(\phi | \data^\text{tr}_\task, \theta)$, where $\theta$ induces a Gaussian prior on the weights with mean $\theta$ and a covariance that depends on the step size and number of gradient steps.\footnote{This result is exact in the case of linear functions, and a local approximation in the case of nonlinear neural networks.} The derivation is outside of the scope of this paper, and we refer the reader to prior work for details~\cite{grant2018recasting,santos}.

In our approach, adaptation involves gradient descent on the learned loss $\loss_\psi(\phi,\data^\text{tr}_\task)$, rather than the likelihood $\log p(\data^\text{tr}_\task | \phi)$. Since we still take a fixed number of steps of gradient descent starting from $\theta$, the result in prior work still implies that we are approximately imposing the Gaussian prior $\log p(\phi |\theta)$~\cite{grant2018recasting,santos}, and therefore are performing approximate MAP inference on the following joint distribution:
\[
p(\phi | \data^\text{tr}_\task, \theta) \propto p(\phi , \data^\text{tr}_\task | \theta) \propto \underbrace{p(\phi | \theta)}_{\text{from GD}} \underbrace{\Psi(\phi, \data^\text{tr}_\task)}_{\exp(-\loss_\psi(\phi,\data^\text{tr}_\task))}.
\]
This is a partially directed factor graph with a learned factor $\Psi$ over $\phi$ and $\data^\text{tr}_\task$ that has the log-energy $\loss_\psi(\phi,\data^\text{tr}_\task)$. Bayesian inference would require integrating out $\phi$, but MAP inference provides a tractable alternative that still produces good results in practice~\cite{grant2018recasting}. Training is performed by directly maximizing $\bcloss(\phi_\task, \data^\text{tr}_\task)$, where $\phi_\task$ is the MAP estimate of $\phi$. Since the behavior cloning loss corresponds to the log likelihood of the actions under a Gaussian mixture policy, we directly train both the prior $\theta$ and the log-energy $\loss_\psi$ such that MAP inference maximizes the log probability of the actions. Note that, since we use MAP inference during training, the model does not necessarily provide well-calibrated probabilities. However, the probabilistic interpretation still helps to shed light on the role of the learned adaptation objective $\loss_\psi$, which is to induce a joint factor on the observations in $\data_\task$ and the policy parameters $\phi$. A visual illustration of the corresponding graphical model is shown in Figure~\ref{fig:graphical_model}.

\section{Network Architectures}
\label{sec:policy}

Now that we have presented our approach, we describe form of the policy $\pi$ and the learned adaptation objective $\loss_\psi$.

\vspace{-0.05cm}
\subsection{Policy Architecture}
\vspace{-0.1cm}

As illustrated in Figure~\ref{fig:policy}, the policy architecture is a convolutional neural network that maps from RGB images to a distribution over actions. The convolutional network begins with a few convolutional layers, which are fed into a channel-wise spatial soft-argmax that extracts 2D feature points $\mathbf{f}$ for each channel of the last convolution layer~\cite{e2e}.
Prior work has shown that the spatial soft-argmax is particularly effective and parameter-efficient for learning to represent the positions of objects in robotics domains~\cite{e2e,dsae}.
Following prior work~\cite{e2e}, we concatenate these feature points with the robot configuration, which consists of the pose of the end-effector represented by the 3D position of 3 non-axis-aligned points on the gripper. Then, we pass the concatenated feature points and robot pose into multiple fully connected layers. The distribution over actions is predicted linearly from the last hidden layer $\mathbf{h}$. We initialize the first convolutional layer from that of a network trained on ImageNet.

In our experiments, we will be using a continuous action space over the linear and angular velocity of the robot's gripper and a discrete action space over the gripper open/close action. Gaussian mixtures can better model multi-modal action distributions compared to Gaussian distributions and has been used in previous imitation learning works~\cite{florida_lstm}. Thus, for the continuous actions, we use a mixture density network~\cite{mdn} to represent the output distribution over actions. For the discrete action of opening or closing the gripper, we use a sigmoid output with a cross-entropy loss.

Following prior work~\cite{vr_imitation}, we additionally have the model predict the pose of the gripper when it contacts the target object and/or container. This is part of the outer meta-objective, and we can easily provide supervision using the robot demonstration. Note that this supervision is not needed at meta-test time when the robot is learning from a video of a human.
For placing and pick-and-place tasks, the target container is located at the final end-effector pose. Thus, we use the last end-effector pose as supervision. For pushing and pick-and-place, the demonstrator manually specifies the time at which the gripper initially contacts the object and the end-effector pose at that time step is used. The model predicts this intermediate gripper pose linearly from the feature points $\mathbf{f}$, and the predicted pose is fed back into the policy.
Further architecture details are included in Section~\ref{sec:experiments}. 

\vspace{-0.05cm}
\subsection{Learned Adaptation Objective Architecture}
\vspace{-0.1cm}

Because we may need to update both the policy's perception and control, the adaptation objective will operate on a concatenation of the predicted feature points, $\mathbf{f}$ (at the end of the perception layers), and the final hidden layer of the policy, $\mathbf{h}$ (at the end of the control layers). This allows the learned loss to more directly adapt the weights in the convolutional layers, bypassing the control layers.
The updated task parameters are computed using our temporal adaptation objective,
$$
\phi = \theta - \alpha \nabla_\theta \loss_{\psi}(\theta, \humandemo),
$$
where we will decompose the objective into two parts: \mbox{$\loss_{\psi} = \loss_{\psi_1}(\mathbf{f}_{1:T}) +  \loss_{\psi_2}(\mathbf{h}_{1:T})$}
We use the same architecture for $\loss_{\psi_1}$ and $\loss_{\psi_2}$, which is illustrated in Figure~\ref{fig:learned_loss}. The learned objective consists of three layers of temporal convolutions, the first two with $10\times1$ filters and the third with $1\times1$ filters. The $\ell_2$ norm of the output of the convolutions is computed to produce the scalar objective value.

\vspace{-0.07cm}
\section{Experiments}
\label{sec:experiments}
\vspace{-0.07cm}

Through our experiments, we aim to address three main questions: (1) Can our approach effectively learn a prior that enables the robot to learn to manipulate new objects after seeing just one video of a human? (2) Can our approach generalize to human demonstrations from a different perspective than the robot, on novel backgrounds, and with new human demonstrators? (3) How does the proposed approach compare to alternative approaches to meta-learning?
In order to further understand our method and its applicability, we additionally evaluate it understand: (a) How important is the temporal adaptation objective? (b) Can our approach be used on more than one robot platform, and with either kinesthetic or teleoperated demonstrations for meta-training?

\begin{figure}
\setlength{\unitlength}{0.5\columnwidth}
\begin{picture}(1.0,0.95) \linethickness{0.5pt}
\includegraphics[width=0.48\columnwidth]{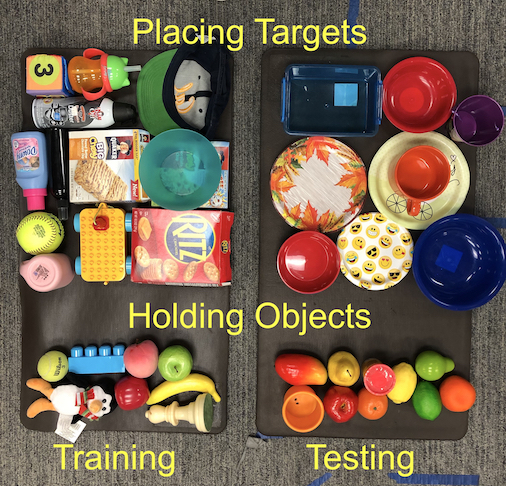}
\includegraphics[width=0.48\columnwidth]{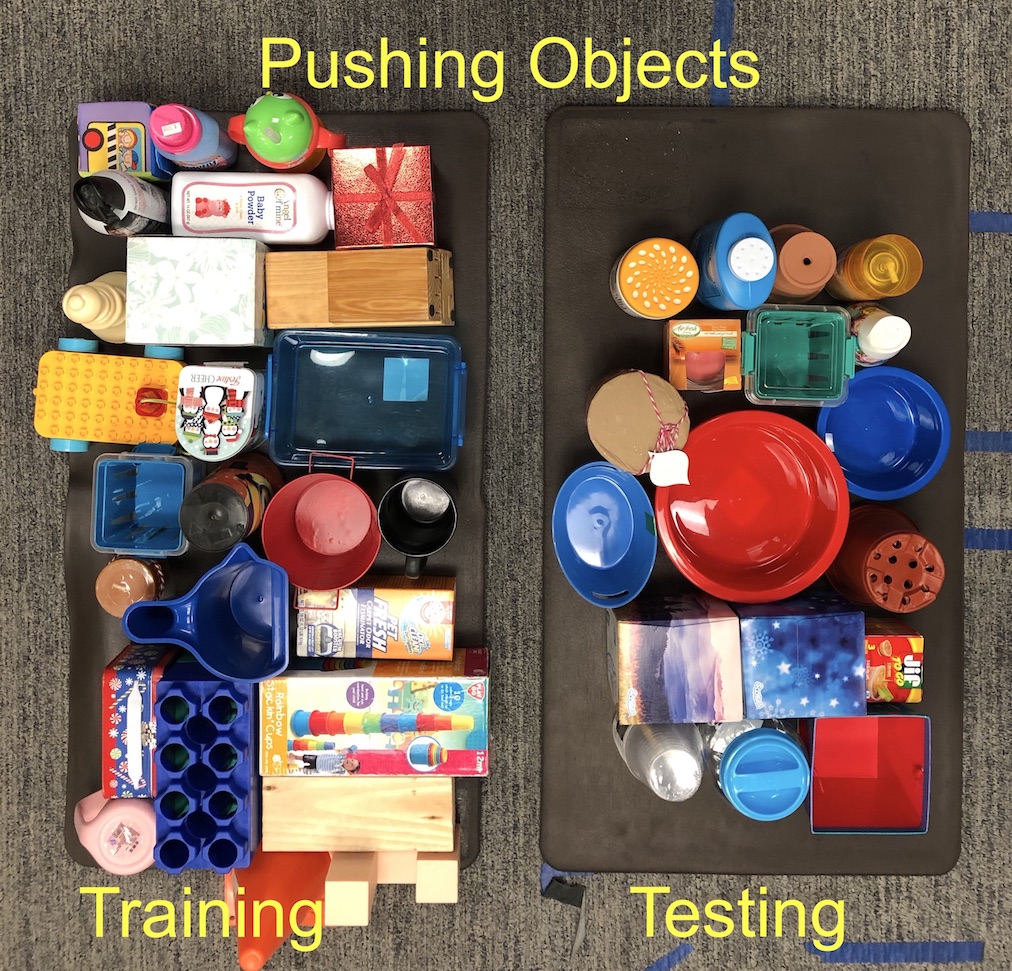}
\end{picture}
\vspace{-0.2cm}
\caption{
\label{fig:objects}
Subset of the objects used for training and evaluation. The robot must learn to recognize and maneuver the novel test objects using just one video of a human.}
\vspace{-0.3cm}
\end{figure}

To answer these questions, we run our experiments primarily with a 7-DoF PR2 arm, with robot demonstrations collected via teleoperation, and RGB images collected from a consumer-grade camera (unless noted otherwise), and use a Sawyer robot with kinesthetic demonstrations to study (b). We compare the following meta-learning approaches:
\vspace{-0.05cm}
\begin{itemize}[leftmargin=*]
    \item \textbf{contextual policy}: a feedforward network that takes as input the robot's observation and the final image of the human demo (to indicate the task), and outputs the predicted action.
    \item \textbf{DA-LSTM policy}: a recurrent network that directly ingests the human demonstration video and the current robot observation, and outputs the predicted robot action. This is a domain-adaptive version of the meta-learning algorithm proposed by~\citet{rocky}.
    \item \textbf{\method, linear loss}: our approach with a linear per-timestep adaptation objective.
    \item \textbf{\method, temporal loss}: our approach with the temporal adaptation objective described in Section~\ref{sec:method}.
\end{itemize}
\vspace{-0.05cm}
All methods use a mixture density network~\cite{mdn} to represent the action space, where the actions correspond to the linear and rotational velocity of the robot gripper, a 6-dimensional continuous action space. As discussed in Section~\ref{sec:policy}, each network is also trained to predict the final end-effector pose using a mean-squared error objective. We train each policy using the Adam optimizer with default hyperparameters~\cite{adam}. All methods use the same data and receive the same supervision. For measuring generalization from meta-training to meta-testing, we use held-out objects in all of our evaluations that were not seeing during meta-training, as illustrated in Figure~\ref{fig:objects}, and new human demonstrators. We provide full experimental details, hyperparameters, and architecture information in Appendix~\ref{app:hyperparameter}. Code for our method will be released upon publication, and we encourage the reader to view the supplementary video.\footnote{The video is available at \url{sites.google.com/view/daml}}

\begin{figure*}
\setlength{\unitlength}{0.5\columnwidth}
\begin{picture}(1.99,0.67) \linethickness{0.5pt}
\put(0,0.0){\includegraphics[width=0.66\columnwidth]{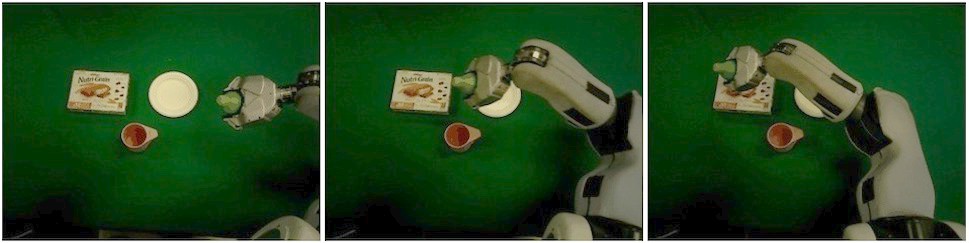}}
\put(0,0.33){\includegraphics[width=0.66\columnwidth]{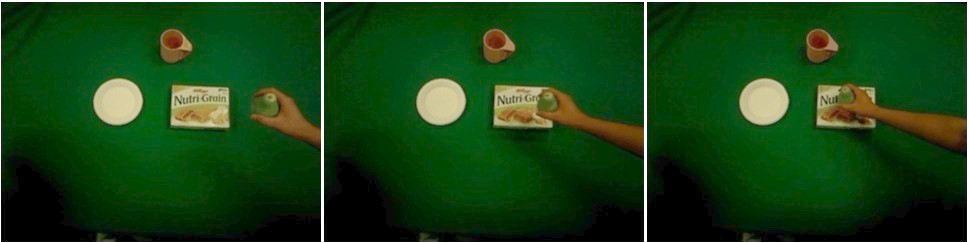}}

\put(1.375,0.0){\includegraphics[width=0.66\columnwidth]{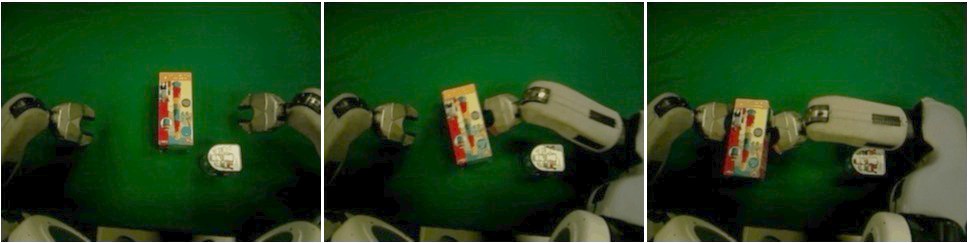}}
\put(1.375,0.33){\includegraphics[width=0.66\columnwidth]{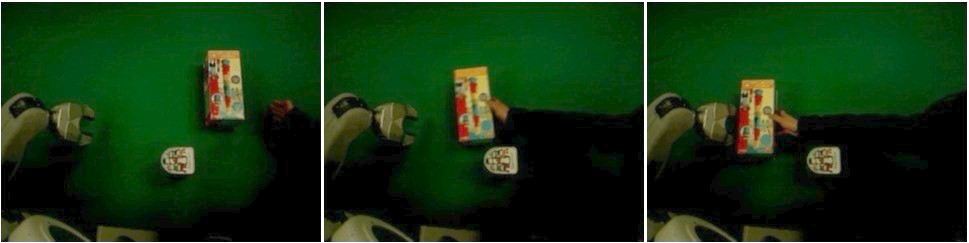}}
\put(2.75,0.0){\includegraphics[width=0.66\columnwidth]{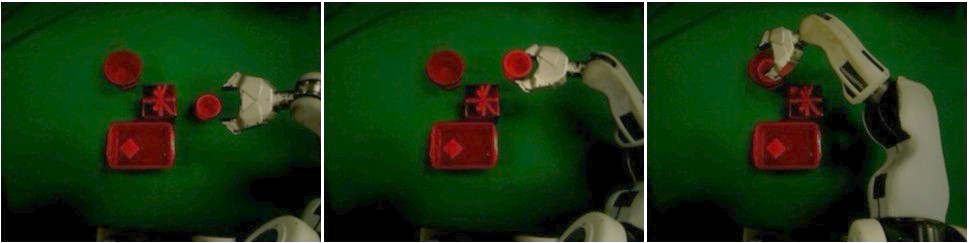}}
\put(2.75,0.33){\includegraphics[width=0.66\columnwidth]{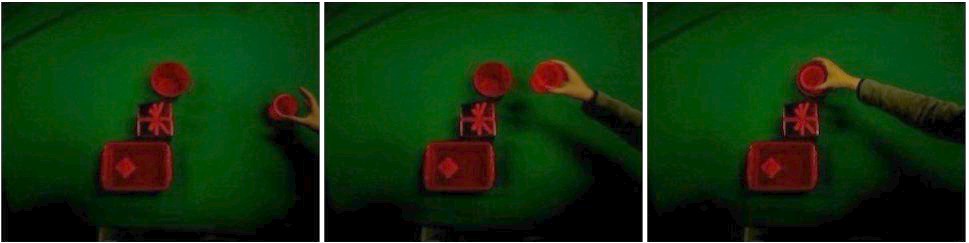}}
\end{picture}
\vspace{-0.25cm}
\caption{
\label{fig:tasks}
Example placing (left), pushing (middle), and pick-and-place (right) tasks, from the robot's perspective. The top row shows the human demonstrations used in Section~\ref{sec:exp1}, while the bottom shows the robot demonstration.}
\vspace{-0.4cm}
\end{figure*}

\begin{figure}
\vspace{0.09cm}
\setlength{\unitlength}{0.5\columnwidth}
\begin{picture}(1.0,0.81) \linethickness{0.5pt}
\put(1.33,0){\includegraphics[width=0.32\columnwidth]{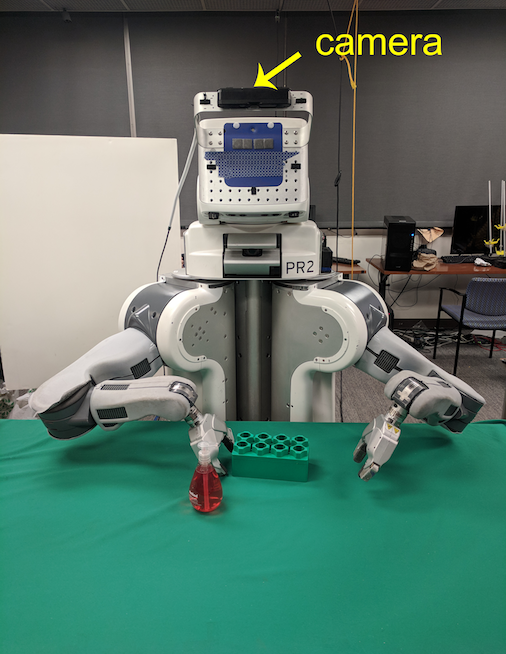}}
\put(0.0,0){\includegraphics[width=0.32\columnwidth]{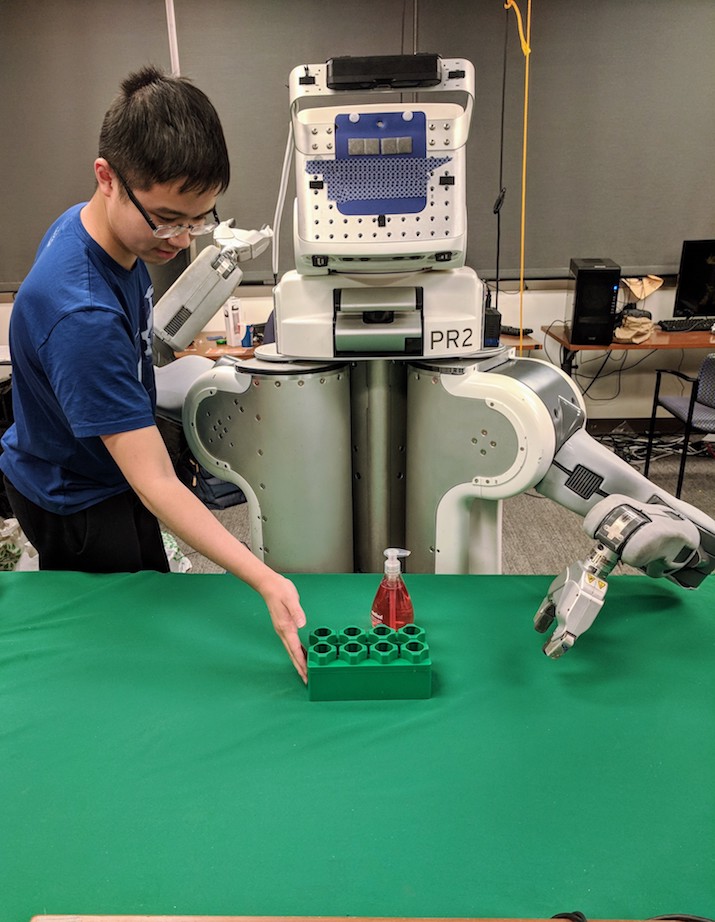}}
\put(0.666,0){\includegraphics[width=0.32\columnwidth]{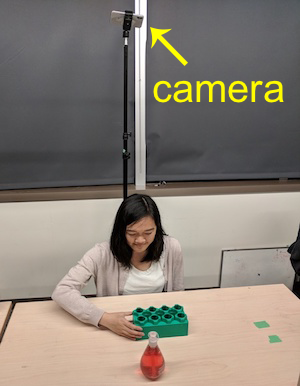}}
\end{picture}
\vspace{-0.2cm}
\caption{
\label{fig:setup}
The PR2 experimental set-up. Left \& Middle: human demonstration set-up from Sections~\ref{sec:exp1} and~\ref{sec:exp2} respectively. Right: test-time set-up.}
\vspace{-0.2cm}
\end{figure}

\vspace{-0.05cm}
\subsection{PR2 Placing, Pushing, and Pick \& Place}
\label{sec:exp1}
\vspace{-0.1cm}

We first consider three different task settings: placing a held object into a container while avoiding two distractor containers, pushing an object amid one distractor, and picking an object and placing it into a target container amid two distractor containers.  The tasks are illustrated in Figure~\ref{fig:tasks}.  In this initial experiment, we collect human demonstrations from the perspective of the robot's camera. For placing and pushing, we only use RGB images, whereas for pick-and-place, RGB-D is used. For meta-training, we collected a dataset with hundreds of objects, consisting of 1293, 640, and 1008 robot demonstrations for placing, pushing, and pick-and-place respectively, and an equal number of human demonstrations.
We use the following metrics to define success for each task: for placing and pick \& place, success if the object landed in or on any part of the correct container; for pushing, success if the correct item was pushed past or within $\sim \!5$ cm of the robot's left gripper.

During evaluation, we used 15, 12, and 15 novel target objects for placing, pushing, and pick-and-place respectively, collected one human demonstration per object, and evaluated three trials of the policy inferred from the human demonstration. We report the results in Table~\ref{tbl:real_results}.  Our results show that, across the board, the robot is able to learn to interact with the novel objects using just one video of a human demo with that object, with pick-and-place being the most difficult task. We find that the DA-LSTM and contextual policies struggle, likely because they require more data to effectively infer the task. This finding is consistent with previous work~\cite{mil}. Our results also indicate the importance of integrating temporal information when observing the human demonstration, as the linear loss performs poorly compared to using a temporal adaptation objective.

\begin{table}[h]
    \begin{center}
    \vspace{-0.2cm}
    \begin{tabular}{|l|c|c|c|}
    \hline
      & placing &  pushing & pick and place\\
     \hline
      DA-LSTM\! &   33.3\% & 33.3\% &  5.6\% \\ 
      \hline
      contextual &   36.1\% & 16.7\% & 16.7\% \\
      \hline
      \method, linear loss   &   76.7\% & 27.8\% & 11.1\% \\
      \hline
      \method, temporal loss (ours) &   \textbf{93.8\%} & \textbf{88.9\%} & \textbf{80.0\%} \\
      \hline
    \end{tabular}
    \end{center}
    \vspace{-0.2cm}
    \caption{\footnotesize One-shot success rate of PR2 robot placing, pushing, and pick-and-place, using human demonstrations from the perspective of the robot. Evaluated using held-out objects and a novel human demonstrator.
    }
    \vspace{-0.8cm}
    \label{tbl:real_results}
\end{table}

\subsection{Demonstrations with Large Domain Shift}
\label{sec:exp2}
\vspace{-0.1cm}

\begin{figure}
\setlength{\unitlength}{0.5\columnwidth}
\begin{picture}(1.0,0.89) \linethickness{0.5pt}
\put(0,0.45){\includegraphics[width=0.245\columnwidth]{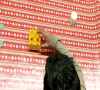}}
\put(0.5,0.45){\includegraphics[width=0.245\columnwidth]{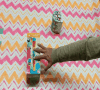}}
\put(1.0,0.45){\includegraphics[width=0.245\columnwidth]{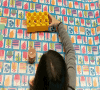}}
\put(1.5,0.45){\includegraphics[width=0.245\columnwidth]{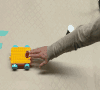}}
\put(0,0){\includegraphics[width=0.245\columnwidth]{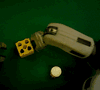}}
\put(0.5,0){\includegraphics[width=0.245\columnwidth]{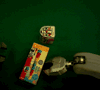}}
\put(1.0,0){\includegraphics[width=0.245\columnwidth]{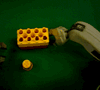}}
\put(1.5,0){\includegraphics[width=0.245\columnwidth]{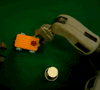}}

\end{picture}
\vspace{-0.2cm}
\caption{
\label{fig:humans}
Human and robot demonstrations used for meta-training for the experiments in Section~\ref{sec:exp2} with large domain shift.
We used ten different diverse backgrounds for collecting human demonstrations. }
\vspace{-0.4cm}
\end{figure}

\begin{figure}
\setlength{\unitlength}{0.5\columnwidth}
\begin{picture}(1.0,0.5) \linethickness{0.5pt}
\put(0.25,0){\includegraphics[width=0.75\columnwidth]{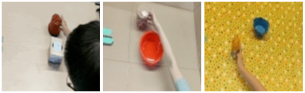}}

\end{picture}
\vspace{-0.2cm}
\caption{
\label{fig:testhumans}
Frames from the human demos used for evaluation in Section~\ref{sec:exp2}, illustrating the background scenes. The leftmost background was in the meta-training set (seen bg), whereas the right two backgrounds are novel (novel bg1 and novel bg2). The objects and human demonstrator are novel.}
\vspace{-0.2cm}
\end{figure}

Now, we consider a challenging setting, where human demonstrations are collected in a different room with a different camera and camera perspective from that of the robot. As a result, the background and lighting vary substantially from the robot's environment. We use a mounted cell-phone camera to record sequences of RGB images on ten different table textures, as illustrated in Figure~\ref{fig:setup}. The corresponding view of the demonstrations is shown in Figure~\ref{fig:humans}. We consider the pushing task, as described in Section~\ref{sec:exp1}, reusing the same robot demonstrations and collecting an equal number of new demonstrations. We evaluate performance on novel objects, a new human demonstrator, and with one seen and two novel backgrounds, as shown in Figure~\ref{fig:testhumans}.

Like the previous experiment, we evaluated with $12$ novel objects and $3$ trials per object. Because we used different object pairs from the previous pushing experiment, the performance is not directly comparable to the results in Table~\ref{tbl:real_results}.
The results for this experiment are summarized in Table~\ref{tbl:gen_results}. As seen in the supplementary video, we find that the robot is able to successfully learn from the demonstrations with a different viewpoint and background. Performance degrades when using a novel background, which causes a varied shift in domain, but the robot is still able to perform the task about 70\% of the time. 
In Table~\ref{tbl:gen_results}, we also include an analysis of the failure modes of our approach, including the number of failures caused by incorrect task identification -- misidentifying the object -- versus control failures -- when the object was clearly correctly identified, but the robot failed to effectively push it. We see that, when the human demonstrations are on a previously seen background, the robot only fails to identify the object once out of 33 trials, whereas failures of this kind are more frequent on the novel backgrounds. Collecting data on a more diverse array of backgrounds, or using some form of background augmentation would likely reduce these types of failures. The number of control failures is similar for all backgrounds, likely indicative of the challenge of physically maneuvering a variety of previously unseen objects.

\begin{table}[h]
    \begin{center}
    \vspace{-0.1cm}
    \begin{tabular}{|l|c|c|c|}
    \hline
    \!\!pushing  &  seen bg & novel bg 1 & novel bg 2 \\
      \hline
      \!\!\method, temporal loss (ours) &  \textbf{81.8\%}  & \textbf{66.7\%}  & \textbf{72.7\%} \\
      \hline
    \end{tabular}
    \end{center}
    \vspace{-0.3cm}
    \begin{center}
    \begin{tabular}{|l|c|c|c|}
    \hline
    \!\!Failure analysis of DAML  &  \!seen bg\! & \!novel bg 1\! & \!novel bg 2\! \\
     \hline
    \!\!\# successes &  27   &  22 & 24 \\ 
      \hline
    \!\!\# failures from task identification  & 1    & 5 &  4\\
      \hline
    \!\!\# failures from control  &  5  & 6  & 5 \\
      \hline
    \end{tabular}
    \end{center}
    \vspace{-0.2cm}
    \caption{\footnotesize Top: One-shot success rate of PR2 robot pushing, using videos of human demonstrations in a different scene and camera, with seen and novel backgrounds. Evaluated using held-out objects and a novel human. Bottom: Breakdown of the failure modes of our approach.
    }
    \label{tbl:gen_results}
    \vspace{-0.9cm}
\end{table}

\vspace{-0.05cm}
\subsection{Sawyer Experiments}
\label{sec:sawyer}
\vspace{-0.1cm}

\begin{figure}
\setlength{\unitlength}{0.5\columnwidth}
\begin{picture}(1.0,0.69) \linethickness{0.5pt}
\put(0.0,0.09){\includegraphics[width=0.29\columnwidth]{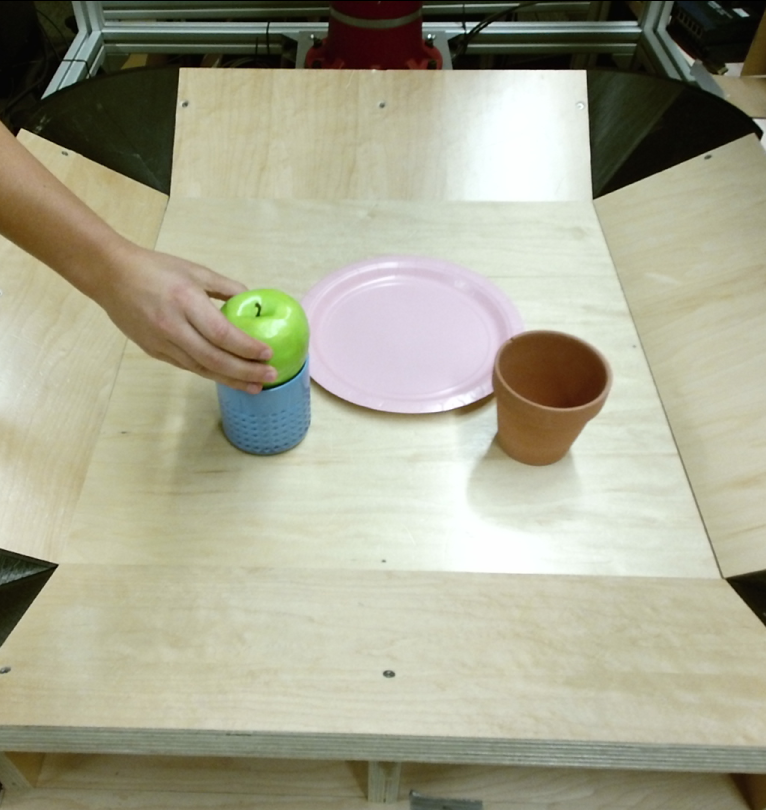}}
\put(0.6,0.09){\includegraphics[width=0.29\columnwidth]{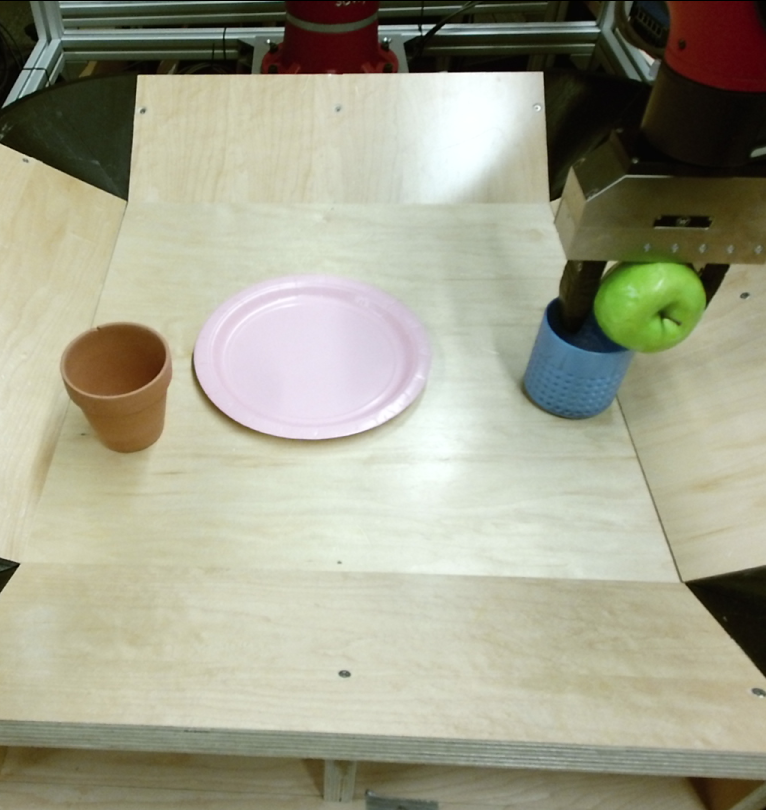}}
\put(1.2,0){\includegraphics[width=0.39\columnwidth]{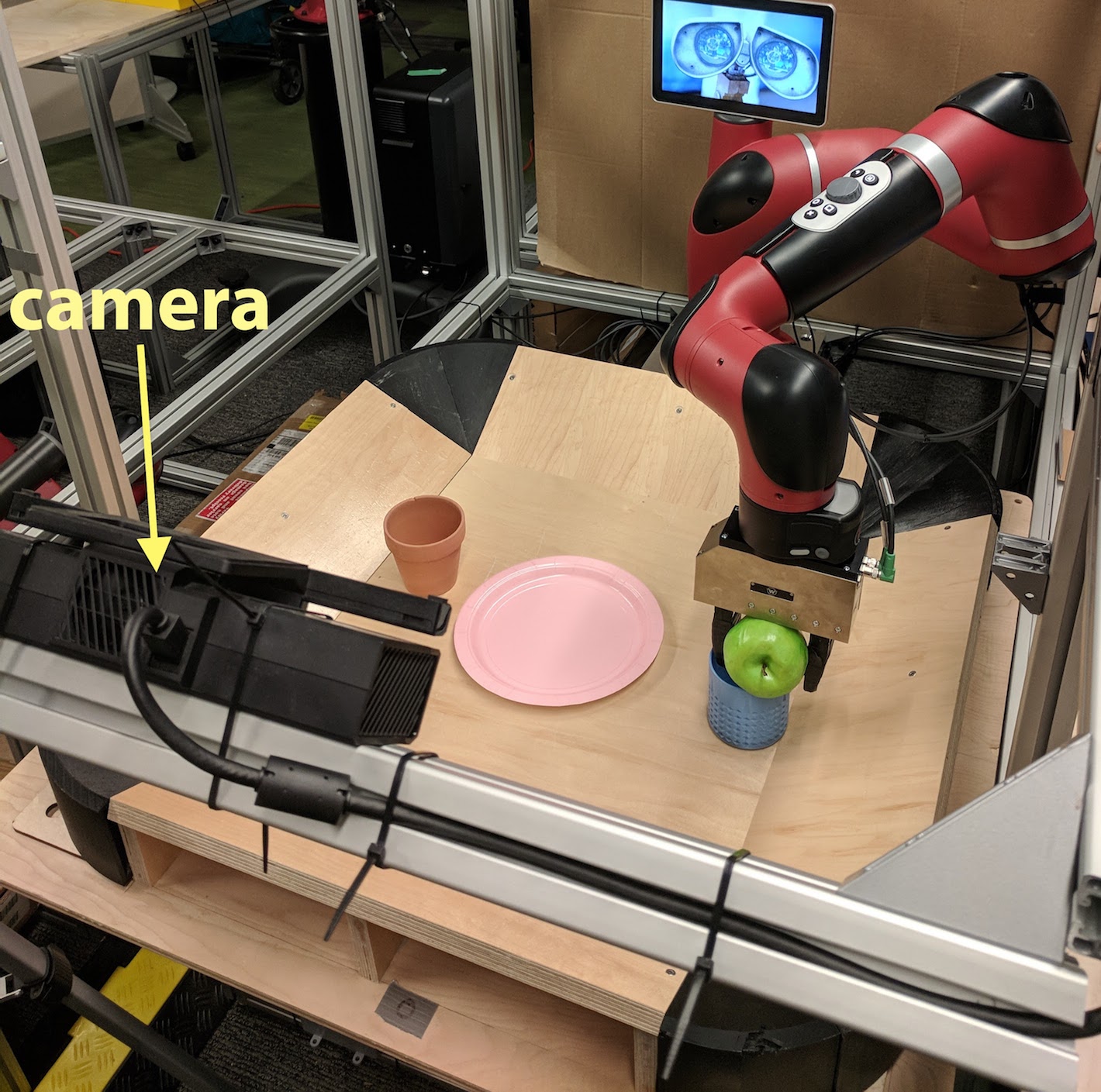}}
\end{picture}
\vspace{-0.2cm}
\caption{
\label{fig:sawyer}
Sawyer robot set-up. From left the right: a human demo from the robot's perspective, the policy execution from the robot's perspective, and an photo illustrating the experimental set-up.}
\end{figure}

The goal of this experiment is to evaluate the generality of our method on a different robot and a different form of robot demonstration collection. We will use a 7-DoF Sawyer robot (see Figure~\ref{fig:sawyer}), and use kinesthetic teaching to record the robot demonstrations, which introduces additional challenges due to the presence of the human demonstrator in the recorded images. The human demonstrations are collected from the perspective of the robot. 
We consider the placing task described in Section~\ref{sec:exp1}. Unlike the PR2 experiments, the action space will be a single commanded pose of the end-effector and we will use mean-squared error for the outer meta-objective.
Since we have thoroughly compared our method on the PR2 benchmarks, we only evaluate our proposed method in this experiment. We evaluated our method using 18 held-out objects and $3$ trials per object. The result was a 77.8\% placing success rate, indicating that our method can successfully be applied to the Sawyer robot and can handle kinesthetic demonstrations during meta-training.

\vspace{-0.05cm}
\subsection{Learned Adaptation Objective Ablation}
\label{sec:simulated}
\vspace{-0.1cm}

Finally, we study the importance of our proposed temporal adaptation objective. In order to isolate just the temporal adaptation loss, we perform this experiment in simulation, in a setting without domain shift. We use the simulated pushing task proposed by~\citet{mil} in the MuJoCo physics engine~\cite{mujoco}. To briefly summarize the experimental set-up, the imitation problem involves controlling a 7-DoF robot arm via torque-control to push one object to a fixed target position amid one distractor, using RGB images as input. The initial positions of the objects are randomized, as is the texture, shape, size, mass, and friction. Meta-training uses 105 object meshes, while 11 held-out meshes and multiple held-out textures are used for meta-testing. A push is considered successful if the target object lands on the target position for at least 10 timesteps within the 100-timestep episode. The results in Table~\ref{table:sim_push} demonstrate a 14\% absolute improvement in success by using a temporal adaptation objective, indicating the importance of integrating temporal information when learning from raw video.

\begin{table}[h]
    \begin{center}
    \vspace{-0.2cm}
    \begin{tabular}{|l|c|}
    \hline
      &  simulated pushing\\ & no domain shift \\
     \hline
      LSTM~\cite{rocky}\! &    34.23\%  \\ 
      \hline
      contextual &     56.98\%  \\
      \hline
      MIL, linear loss~\cite{mil}  &  66.44\%    \\
      \hline
      MIL, temporal loss (ours) &   \textbf{80.63\%}   \\
      \hline
    \end{tabular}
    \end{center}
    \vspace{-0.2cm}
    \caption{\footnotesize One-shot success rate of simulated 7-DoF pushing using video demonstrations with no domain shift
    }
    \vspace{-0.8cm}
    \label{table:sim_push}
\end{table}

\vspace{-0.05cm}
\section{Discussion} 
\label{sec:conclusion}
\vspace{-0.1cm}

We presented an approach that enables a robot learning to visually recognize and manipulate a new object after observing just one video demonstration from a human user. To enable this, our method uses a meta-training phase where it acquires a rich prior over human imitation, using both human and robot demonstrations involving other objects. Our method extends a prior meta-learning approach to allow for learning cross-domain correspondences and includes a temporal adaptation loss function.
Our experiments demonstrate that, after meta-learning, robots can acquire vision-based manipulation skills for a new object using from video of a human demonstrator in a substantially different setting.

\noindent \textbf{Limitations}: While our work enables one-shot learning for manipulating new objects from one video of a human, our current experiments do not yet demonstrate the ability to learn entirely new motions in one shot. The behaviors at meta-test time are structurally similar to those observed at meta-training time, though they may involve previously unseen objects and demonstrators. We expect that more data and a higher-capacity model would likely help enable such an extension. However, we leave this to future work. An additional challenge with our approach is the amount of demonstration data that is needed for meta-training. In our experiments, we used a few thousand demonstrations from robots and humans. However, the total amount of data \emph{per-object} is quite low (around 10 trials), which is one or two orders of magnitude less than the number of demonstrations per-object used in recent single-task imitation learning works~\cite{florida_lstm,vr_imitation}. Thus, if the goal is to enable a \emph{generalist} robot that can adapt to a diverse range of objects, then our approach is substantially more practical. 

\noindent \textbf{Beyond Human Imitation:} While our experiments focus on imitating humans, the proposed method is not specific to perceiving humans, and could also be used, for example, for imitating animals or a simulated robot, for simulation to real world transfer. 
Beyond imitation, we believe our approach is likely more broadly applicable to problems that involve inferring information from out-of-domain data, such as one-shot object recognition from product images, a problem faced by teams in the Amazon Robotics Challenge~\cite{zeng2018robotic}.

\section*{Acknowledgments}
We thank Saurabh Gupta, Rowan McAllister, and Dinesh Jayaraman for helpful feedback on an early version of this paper.

{\small
\bibliographystyle{abbrvnat}
\bibliography{references}
}
\clearpage

\appendix
\section{Hyperparameter and Experimental Details}
\label{app:hyperparameter}

In this appendix, we include experimental details and hyperparameters for each of the experiments.

All of the human and robot demonstrations are collecting at 10 Hz, and take approximately 3-8 seconds. Then, for meta-training, we randomly sampled 40 of the frames and corresponding actions to be used as the demonstration.
The robot policy is executed at 10 Hz.

\subsection{Placing, Pushing, and Pick-and-Place Experiments}
\label{app:exp1}

We include the details of the experiments in Section~\ref{sec:exp1} for the placing, pushing, and pick-and-place tasks.
The architecture and hyperparameters were selected by evaluating the end-effector loss and control loss on a validation set of $12$, $8$, $12$ objects for placing, pushing, and pick-and-place respectively, sampled and held out from the training data. For placing and pushing, the inputs are RGB images with size $100 \times 90$, and for pick-and-place, we use RGB-D images with size $110 \times 90$. We train separate models for each of the three settings. 

For placing, the policy architecture uses $5$ convolutional layers with $64$ $3\times 3$ convolutional filters in each layer where the first three layers are with stride $2$ and the last two layer is with stride $1$. The first convolutional layer uses pretrained weights from VGG-$19$. It also uses $3$ fully connected layers of size $100$, and a learned adaptation objective with two layers of $32$ $10 \times 1$ convolutional filters followed by one layer of $1\times 1$ convolutions. For pushing, the policy architecture, uses the same convolutional network, $4$ fully connected layers of size $50$, and a learned adaptation objective with two layers of $10$ and $20$ $10\times 1$ convolutional filters for action and final gripper pose prediction respectively followed by one layer of $1\times 1$ convolutions. For pick-and-place, the policy architecture uses the same convolutional network except that the first convolutional layer is not pretrained. It also uses $16$ $3 \times 3$ convolutional filters that operate on the depth input and concatenates the depth stream to the RGB stream after the first convolutional layer, channel-wise, following the approach by \citet{vr_imitation}. 
For pick-and-place, we use two gripper poses -- one intermediate, when the gripper contacts the item to pick, and one final pose at the end of the trajectory.
The architecture also uses $4$ fully connected layers of size $50$, and a learned adaptation objective with two layers of $10$, $30$, and $30$, $10\times 1$ convolutional filters for action, final gripper pose and pickup gripper pose prediction respectively followed by one layer of $1\times 1$ convolutions.
All architectures use ReLU nonlinearities and layer normalization. 

All the baseline methods use the same architecture for convolutional layers as DAML for each experiment. DAML with a linear adaptation objective also uses the same fully-connected architecture as DAML with the temporal adaptation objective except that its learned adaptation objective consists of one linear layer. The LSTM uses $512$ LSTM hidden units for all experiments. The contextual model uses $3$ fully connected layers with size $100$ for all experiments. 

For placing, we use a behavioral cloning loss as a combination of $\ell_1$ and $\ell_2$ losses, where the $\ell_2$ loss is scaled down by a factor of $100$, following prior work~\cite{mil}. For the pushing and pick-and-place experiments, all methods use a mixture density network as mentioned in Section~\ref{sec:policy} after the last fully-connected layer with 20 modes and the negative log likelihood of the mixture density network as the behavioral cloning loss. At test time, at each time step, we sample $100$ actions from the learned mixture distribution and choose the action with highest probability. For DAML with both a linear and temporal adaptation objective, we use a step size $\alpha=0.01$ for placing and $\alpha = 0.005$ for pushing and pick-and-place with inner gradient clipping within the range $[-30, 30]$. We use $12, 10$, and $4$ tasks in the meta batch at each iteration for placing, pushing and pick-and-place respectively. For all methods, we use $1$ human demonstration and $1$ robot demonstration for each sampled task. We train the model for $50$k iterations for placing and placing, and $75$k iterations for pick-and-place. We use $5$ inner gradient update steps and a bias transformation with dimension $20$ for all experiments. Since we don't have the robot state $\mathbf{s}$ for human demonstrations, we set the state input to be $0$ when computing inner gradient update and feed the robot states into the policy when we update the policy parameters with robot demonstrations.


\subsection{Diverse Human Demonstration Experiments}
\label{app:exp2}

We include the details of the experiments in Section~\ref{sec:exp2} using diverse human demonstrations. For meta-training, eight pushing demonstrations are taken for each of $80$ total objects grouped into $40$ pairs. Each demo is shot in front of a randomly selected  background, among 10 backgrounds. The viewpoint for the human demos is held fixed with a phone camera mounted on a tripod. Before being fed into the model during training images are modified with noise sampled uniformly from the range $[-0.3, 0.3]$ to their lighting. This color augmentation process helps the model perform more robustly in different light conditions. We use the same input image size and policy architecture as the pushing experiment described in Section~\ref{sec:exp1} and Appendix~\ref{app:exp1}.

\subsection{Sawyer Robot Experiments}

We include the details of the experiments in Section~\ref{sec:sawyer} on the Sawyer robot. The primary differences between this experiment and previous placing experiments are the robot used and how demonstrations were collected. While PR2 robot demonstrations were taken using a teleoperation interface, the Sawyer arm was controlled kinesthetically by humans: the demonstrator guided the Sawyer arm to perform the goal action. Demonstrations were collected at 10 Hz. Saved at each timestep are a monocular RGB image taken by a Kinect sensor, the robot's joint angles, its joint velocities, and its gripper pose. 

The architecture and hyperparameters were tuned by evaluating the gripper pose loss on a held out validation set of 20 objects. 
The policy architecture takes in RGB images of size $100 \times 100$. It uses the same convolutional and fully-connected layers as well as squared error behavioral cloning loss as in the model for the PR2 placing experiment, and a learned adaptation objective with three layers of $32$ $20 \times 1$ convolutional filters followed by one layer of $1\times 1$ convolutions. We use a step size $\alpha=0.005$ with inner gradient clipping within the range $[-30, 30]$, $8$ as the meta batch size, and $1$ human demonstration as well as $1$ robot demonstration for each sampled task. We train the model for $60$k iterations.

During training, we augmented the images using random color augmentation, by adding noise uniformly sampled in $[-0.3, 0.3]$ to their hue, saturation, and value.
The images used during evaluation were not modified. To control the robot during evaluation, the first image frame is used to predict the final end-effector pose of the robot. After the robot reaches the predicted gripper pose, the robot is controlled using the prediction actions, which are continuous end-effector velocities. At this point the gripper is opened and the robot drops the held item (hopefully) into the target container.

\subsection{Simulated Pushing Experiment}

Here, we include details on the experiments in Section~\ref{sec:simulated}. The pushing environment was introduced and open-sourced by~\citet{mil}.
The expert policy for collecting demonstrations was computed using reinforcement learning. 
Following~\cite{mil}, we compute the reported success rates over 74 tasks with 6 trials per task, totalling to 444 trials. The time horizon is $T=100$. A trial is considered successful if the target object lands on the target position for at least 10 timesteps within the 100-timestep episode.

The inputs are RGB images of size $125 \times 125$. The policy architecture uses $3$ convolutional layers with $16$ $5 \times 5$ filters and stride $2$ followed by $1$ convolutional layer with $32$ $5 \times 5$ filters with stride $1$. It also has $2$ fully-connected layers of size $400$, and a learned adaptation objective with two layers of $64$ $10\times 1$ convolutional filters followed by one layer of $1\times 1$ convolutions. The policy operates on a $125 \times 125$ RGB image, along with the robot joint angles, joint velocities, and end-effector pose. The behavioral cloning loss is the mean squared error between the predicted actions and the ground truth robot commands. We use step size $\alpha=0.01$ with inner gradient clipping within the range $[-20, 20]$, and one inner gradient update step. We use $15$ as the meta batch size, and $2$ different robot demonstrations for each sampled task. We train our policy for $30$k iterations.

\end{document}